\documentclass[pmlr]{jmlr}% new name PMLR (Proceedings of Machine Learning)

\RequirePackage{graphicx}
\usepackage{wrapfig}
\usepackage{lastpage}
 \usepackage{multirow}
 \usepackage{booktabs}
 \usepackage{listings}
 \usepackage{comment}
 \usepackage{placeins} 
\usepackage{longtable}% for long tables
  \usepackage[utf8]{inputenc}    % safe with pdfLaTeX
\usepackage[T1]{fontenc}

\DeclareUnicodeCharacter{202F}{\,}     % narrow no-break space → thin space
\DeclareUnicodeCharacter{00A0}{~}      % NBSP → nonbreaking space
\DeclareUnicodeCharacter{2212}{-}      % minus sign → hyphen-minus

\makeatletter
\def\set@curr@file#1{\def\@curr@file{#1}} %temp workaround for 2019 latex release
\makeatother
\usepackage[load-configurations=version-1]{siunitx} % newer version

\theorembodyfont{\upshape}
\theoremheaderfont{\scshape}
\theorempostheader{:}
\theoremsep{\newline}

\jmlrvolume{298}
\jmlrpages{\pageref{jmlrstart}--\pageref{LastPage}}
\jmlryear{2025}
\jmlrworkshop{Machine Learning for Healthcare}

\title[Improving ARDS Diagnosis Through Context-Aware Concept Bottleneck Models]{Improving ARDS Diagnosis Through Context-Aware Concept Bottleneck Models}

\author{\Name{Anish Narain}
       \Email{anish.narain21@imperial.ac.uk}\\ 
       \addr Imperial College London
       \AND
       \Name{Ritam Majumdar}
    \Email{r.majumdar24@imperial.ac.uk}\\ 
       \addr Imperial College London
        \AND
       \Name{Nikita Narayanan}
\Email{nikita.narayanan18@imperial.ac.uk}\\ 
       \addr Imperial College London
        \AND
       \Name{Dominic Marshall}
    \Email{dominic.marshall12@imperial.ac.uk}\\ 
       \addr Imperial College London 
       \AND
\Name{Sonali Parbhoo}
\Email{s.parbhoo@imperial.ac.uk}\\ 
       \addr Imperial College London
}

\begin{document}

\maketitle

\begin{abstract}
 Large, publicly available clinical datasets have emerged as a novel resource for understanding disease heterogeneity and to explore personalization of therapy. These datasets are derived from data not originally collected for research purposes and, as a result, are often incomplete and lack critical labels. Many AI tools have been developed to retrospectively label these datasets, such as by performing disease classification; however, they often suffer from limited interpretability. Previous work has attempted to explain predictions using Concept Bottleneck Models (CBMs), which learn interpretable concepts that map to higher-level clinical ideas, facilitating human evaluation. However, these models often experience performance limitations when the concepts fail to adequately explain or characterize the task. We use the identification of Acute Respiratory Distress Syndrome (ARDS) as a challenging test case to demonstrate the value of incorporating contextual information from clinical notes to improve CBM performance. Our approach leverages a Large Language Model (LLM) to process clinical notes and generate additional concepts, resulting in a 10\% performance gain over existing methods. Additionally, it facilitates the learning of more comprehensive concepts, thereby reducing the risk of information leakage and reliance on spurious shortcuts, thus improving the characterization of ARDS.\footnote{The code is publicly available at \url{https://github.com/ai4ai-lab/context-aware-cbms}}
\end{abstract}

\section{Introduction}
Retrospective identification of Acute Respiratory Distress Syndrome (ARDS) remains a persistent diagnostic challenge in critical care medicine. If cases of ARDS can be accurately identified from large-scale, routinely collected electronic health records (EHR), this would not only yield valuable epidemiological insights but also offer a rich resource for studying disease heterogeneity and treatment response. ARDS is under-recognized, under-documented and thus poorly coded \citep{Herasevich2009, Bechel2023, Poulose2009}. Accurate identification of ARDS is essential in retrospective studies aimed at evaluating interventions and outcomes. Automated and retrospective identification of ARDS is challenging as it requires integration of structured tabular data with subjectively interpreted clinical information, such as chest radiographs and unstructured free-text clinical notes \citep{Rubulotta2024, ranieri2012acute}. The current gold standard involves expert case review which is costly and time consuming \citep{Rubulotta2024}. While machine learning (ML) models have shown potential for automating ARDS detection from structured data \citep{le2020supervised, zeiberg2019machine}, they often fail to capture the full clinical reasoning process and typically offer limited interpretability—posing a major barrier to adoption in healthcare settings.  

To address these concerns, Concept Bottleneck Models (CBMs) have been proposed as an interpretable alternative to end-to-end black-box models \citep{koh2020concept}. CBMs decompose prediction into two stages: first, the model predicts a set of predefined, human-interpretable concepts from input features; then, these concepts are used to make the final prediction. This architecture enables transparency and the possibility of clinician intervention at the concept level: an important property in safety-critical domains like healthcare. However, despite their promise, CBMs face several limitations that hinder their practical deployment in complex clinical tasks.

A central issue in CBMs is \emph{concept leakage}, where the model learns to infer intermediate concepts using information that is statistically dependent on the target labels, rather than purely from the input features \citep{mahinpei2021promises}. This causes the learned concept distribution $p(c|x)$ to become entangled with the label distribution $p(y|x)$ , such that concept predictions inadvertently reflect label information. When this happens, the CBM's performance is artificially inflated during training, but the model fails to generalize to out-of-distribution or real-world settings where these dependencies do not hold. In the clinical context, structured concepts such as ``$PaO_2/FiO_2 < 300$'' may be defined based on, or exhibit strong correlation with, ARDS labels, making them especially susceptible to information leakage. %SP: TO DO: edit this example

To mitigate these limitations, we propose a hybrid, context-aware Concept Bottleneck Model that integrates structured EHR data with \emph{LLM-derived concepts} from unstructured clinical notes to improve retrospective diagnosis of ARDS. LLM-derived concepts are generated from sources such as radiology reports, discharge summaries, and echocardiography interpretations, which are authored independently of the labeling process and often reflect rich clinical context. Because LLMs infer these representations directly from descriptive documentation rather than structured variables aligned with labels, the resulting concept distribution $ p(c_{\text{text}}|x_{\text{text}})$ is less likely to be conditionally dependent on the outcome $y$ given the input, reducing risk of leakage. This additional information helps the model learn more robust and faithful intermediate representations—ones that are informative, interpretable, and not artificially linked to the target label. Our approach trains a standard CBM on structured clinical variables to capture core physiological features. In parallel, we use a large language model to extract contextual concepts from unstructured notes. These concepts are integrated into the CBM’s bottleneck layer, forming a multi-modal representation that combines the structure of traditional CBMs with the contextual richness of clinical text.

Our contributions are as follows: (i)  We propose a general framework for augmenting CBMs with context from unstructured data, which can be applied to other clinical use cases requiring multi-modal reasoning. (ii) We demonstrate improved retrospective identification of ARDS using a real-world ICU dataset, showing gains in predictive performance by 8-10\%. (iii) We show that augmenting the concepts using LLM-derived contextual concepts improves the completeness of the concept-space. This reduces the model’s reliance on structured variables that may encode spurious correlations with the label, thereby mitigating concept leakage and improving the mutual information between concepts and outcomes. (iv) Finally, our model enables  transparent, concept-level reasoning, allowing for interventions on misclassified patients, erroneous concepts or shortcut-induced errors. We demonstrate that targeted corrections of mislabeled or erroneous concepts can recover misclassified cases, further improving concept-label alignment and fostering more reliable model behavior in deployment. 

\begin{figure}[h]
    \centering
    \includegraphics[scale=0.31]{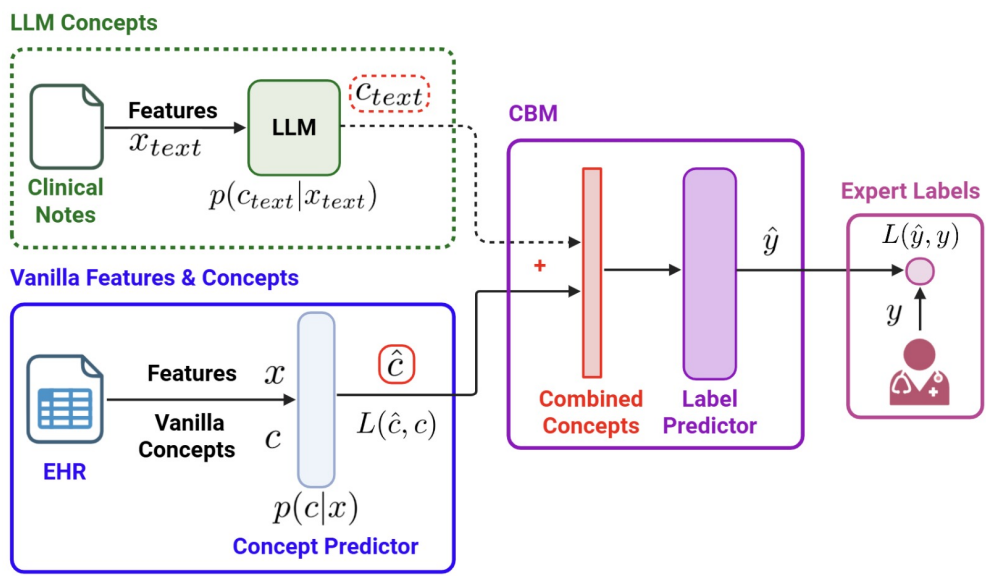}
    \caption{Training pipeline for Context-Aware CBMs. The Vanilla Features \& Concepts Box contains the concept predictor, which learns the concept distribution $p(c|x)$ from structured EHR data. We use an LLM (top left) to extract concepts from the unstructured clinical notes, resulting in a separate distribution $ p(c_{\text{text}}|x_{\text{text}})$, which differs from $p(c|x)$. These two concept distributions improve the completeness of the concept space and are concatenated to predict the final label $y$. }
    \label{fig:pipeline}
\end{figure}
\subsection*{Generalizable Insights about Machine Learning in the Context of Healthcare}
Our work highlights several broader lessons for the design and deployment of machine learning models in clinical settings, particularly when interpretability and data heterogeneity are key concerns. Although our case study focuses on the retrospective diagnosis of ARDS, it offers broader lessons that are generalizable to a wide range of clinical applications and inform the development of more reliable and actionable ML systems. Specifically,

\begin{itemize}
\item \emph{Unstructured Text Offers a Natural Regularizer for Representation Learning.} Text-derived concepts, when generated from physician-authored notes, provide an alternative view of the patient state that is often less correlated with labeling artifacts. These representations, extracted via LLMs, act as a form of distributional regularization — encouraging the model to learn more generalizable and semantically grounded features. That is, LLMs can serve not just as predictors, but as tools for distilling rich, weakly supervised signals that regularize downstream models.

\item \emph{Shortcut Learning Can Emerge Even in Human-Defined Features.}
Shortcut learning is not limited to raw inputs—our results show that structured, curated clinical variables can also act as shortcuts if they reflect or proxy the label. This extends shortcut risk to features typically seen as safe or interpretable, highlighting the need for diverse, independent data to avoid label-dependent overfitting.

\item \emph{Multi-Modal Models Reflect Human-Like Reasoning.}
Our model integrates structured and unstructured data to reflect how clinicians combine quantitative metrics with qualitative narratives, improving both predictive robustness and alignment with human reasoning—especially for complex, poorly codified conditions. Multi-modal integration enhances not just performance, but also clinical interpretability.

\end{itemize}

\section{Related Work}
\paragraph{Concept Leakage and Shortcut Learning.}
Several works highlight the problem of concept leakage in CBMs where the models exploit unintended information to improve predictive performance, e.g. \citep{havasi2022addressing,margeloiu2021concept}. In the same vein, several approaches have been developed to address shortcut learning in deep models \citep{geirhos2020shortcut,makar2022causally}. Some of these methods employ regularization to discourage shortcut variables from influencing predictions \citep{wang2022learning,makar2022causally}. But regularization methods often suppress not only spurious features but also causally relevant ones, reducing overall performance \citep{hong2025on}. Others have shown that augmenting a dataset with additional information can help decorrelate shortcuts from data \citep{cubuk2019autoaugment}. We build on this idea to show how augmenting concept-based explanations with additional contextual information from unstructured clinical notes using LLMs can reduce a model's reliance on shortcuts and concept leakage overall.

\paragraph{Clinical Concept Bottleneck Models.} 
CBMs are supervised learning models that map raw inputs $x$ to human-understandable concepts $c$, and then to target labels $y$, offering insights into the model's reasoning (\cite{margeloiu2021concept}). CBMs have been applied across a range of clinical tasks, including predicting vasopressor onset from structured EHR data \citep{wu2022learning}, forecasting ICU interventions \citep{ghassemi2017predicting}, and classifying medical images using interpretable visual markers \citep{laguna2024beyond, yan2023robust}. \citet{koh2020concept} showed that CBMs could identify arthritis by first predicting radiographic features like joint space narrowing. While these models improve interpretability, they often rely on predefined structured features and lack access to the rich contextual information present in clinical narratives. Our work addresses this gap by augmenting CBMs with additional concepts extracted from unstructured clinical notes using LLMs, enabling improved performance on complex retrospective tasks such as ARDS diagnosis.

\paragraph{Clinical Large Language Models, Attention, and Transformers.}
Attention-based models have been widely employed in healthcare to enhance interpretability by assigning weights to input features, thereby highlighting their contributions to predictive outcomes \citep{de2021enhancing,chen2020interpretable}. However, these models primarily focus on structured data and may not fully capture the rich contextual information present in unstructured clinical notes as we propose in our work. Recent advancements in LLMs have shown promise in clinical settings by extracting insights from unstructured clinical text, such as radiology reports, discharge summaries, and clinical notes. For instance, MedPaLM-2 and Llama-3 have excelled on medical exams  \citep{singhal2023towards, zhou2023survey} and BioBERT has been successful in extracting clinical concepts and predicting patient outcomes from free-text data \citep{lee2020biobert}. Although these models provide powerful insights into unstructured data, they often face limitations such as poor performance in complex cases (\cite{chen2024benchmarking}), costly updates, and lack of interpretability, which limits their utility in clinical settings. We address these gaps by using LLMs for concept generation within a CBM, leveraging the interpretability of the models with the text-processing strengths of LLMs. There has also been work to use transformers to predict diseases directly from clinical notes, but this presents key challenges. Standard transformers struggle with long, dispersed documentation due to fixed input lengths (\cite{islam2025llm}), leading to truncation-related loss, high computational cost (\cite{vaswani2017attention}), and limited ability to capture long-range dependencies (\cite{zhang2025leveraging}). Signals are often diluted by routine content, making raw note processing noisy and inefficient. Our method prompts an LLM to extract predefined, clinically meaningful concepts from unstructured text, yielding interpretable features that retain signal while filtering noise. This reduces complexity for downstream CBMs, improves model focus, and avoids black-box issues in raw-text models.

\paragraph{Methods for ARDS Identification.} 
Current machine learning approaches for identifying ARDS have improved early detection but often lack specificity and generalization. Many models, known as ``sniffer systems" (\cite{wayne2019electronic}), rely on structured EHR data like vital signs and radiology reports but often overfit and fail to generalize across datasets (\cite{mckown2019external}). Some use keyword searches in radiology notes (\cite{afshar2018computable}), but this lacks adaptability due to institutional variations in terminology. Recent efforts using LLM-based models (\cite{pathak2023respbert}) show promise in improving generalization by analyzing clinical notes without predefined keywords. Unlike these methods, we leverage the performance of LLMs for text-data and CBMs to deduce a set of interpretable concepts that enable more accurate diagnosis and characterization of ARDS.

\section{Cohort} \label{sec:cohort}
 \paragraph{Cohort Selection.} 
We curated a balanced cohort of ARDS and non-ARDS patients from the MIMIC-IV database \citep{johnson2023mimic}. As with other large critical care datasets, MIMIC-IV lacks an explicit ARDS label. To define the cohort, we selected adult patients ($\geq$ 18 years) with respiratory failure with the following inclusion criteria: PF ratio $<$ 300 with PEEP $\geq$ 5 cm H$_2$O for three consecutive days, or two days if the patient died on day 3. Exclusion criteria included pregnancy and use of extracorporeal membrane oxygenation (ECMO).  These criteria extend prior work, which often used less rigorous methods such as physiological thresholds combined with keyword searches in radiology reports (\cite{afshar2018computable}, \cite{gandomi2022ardsflag}). Initial validation of those approaches revealed frequent misclassification—especially of patients with transient hypoxia—and failed to exclude cardiac or fluid overload causes of respiratory failure. Our stricter criteria ensure a more reliable enrichment for ARDS before classification. An expert clinician then reviewed clinical records (free-text notes, radiology reports, and echocardiograms) and assigned each patient a label. Of the 1,953 patients reviewed, 1,030 (52.7\%) were labeled positive for ARDS, and 923 (47.3\%) were labeled negative. Table \ref{tab:cohort-stats} summarizes cohort statistics.

\begin{table}[ht]
\centering
\begin{tabular}{lll}
\toprule
\textbf{Variable} & \textbf{Full Dataset} & \textbf{Unseen Data} \\
\midrule
n & 1953 & 50 \\
ARDS Positive & 1030 (52.7\%) & 21 (42\%) \\
ARDS Negative & 923 (47.3\%) & 29 (58\%) \\
\midrule
Female & 732 (37.5\%) & 18 (36\%) \\
Male & 1221 (62.5\%) & 32 (64\%) \\
\midrule
Avg time in ICU (median) & 13.6 days (10.8 days) & 13.73 days (10 days) \\
Age ($\mu \pm \sigma$) & 62.6 $\pm$ 15.4 & 65.1 $\pm$ 11.6 \\
\midrule
\textbf{Ethnicity} & & \\
\midrule
White & 1261 (64.6\%) & 30 (55.6\%) \\
Black or African-American & 147 (7.5\%) & 6 (11.1\%) \\
Hispanic or Latino & 75 (3.8\%) & 1 (1.9\%) \\
Other & 131 (6.7\%) & 6 (11.1\%) \\
Not Available & 339 (17.4\%) & 11 (20.4\%) \\
\bottomrule
\end{tabular}
\caption{ARDS cohort statistics—comparison between full dataset and unseen data.}
\label{tab:cohort-stats}
\end{table}

\paragraph{Data Preprocessing and Feature Extraction.} 
We curated structured and unstructured EHR data from 1,953 patients in the MIMIC-IV database using subject, stay, and admission identifiers. For structured data, we selected features reflecting both systemic organ dysfunction and respiratory status, in line with clinical characteristics of ARDS. These included components of the SOFA score (respiratory, cardiovascular, renal, central nervous system), vasopressor support (norepinephrine equivalent dose), mechanical ventilation duration, and pre-existing respiratory comorbidities (grouped into six diagnostic categories using ICD-9/10 codes). For each SOFA component, we extracted relevant lab values corresponding to the worst hourly value across the stay, the stay-level average, and the worst value in the first 24 hours. Vasopressor exposure was captured using the time-weighted average and peak norepinephrine dose. Mechanical ventilation duration was calculated across the entire stay. All continuous features were scaled to [0, 1] using min-max normalization, and features with $>50\%$ missingness were dropped. Remaining missing values were imputed using the median-value imputation, which is more robust to outliers. The full list of features and concepts can be found in Appendix \ref{sec:dataset}.

\paragraph{Defining Vanilla Concepts from Tabular EHR Data.} We defined a set of vanilla concepts based on the tabular EHR data. These concepts will be used in conjunction with context-specific concepts derived from free-text data to inform predictions in the next section. The vanilla concepts capture variation in SOFA scores and pre-existing respiratory conditions. SOFA-based concepts included organ-specific SOFA scores as well as the worst hourly SOFA score across a patient's stay, the average SOFA score across the stay and the worst hourly sofa score within the first 24 hours. Concepts based on pre-existing respiratory conditions were derived directly on the basis of condition severity: moderate (1-2 pre-existing conditions) and severe (3+ pre-existing conditions). These concepts are binary. Like the feature set, any missing concept values were imputed using median value imputation and non-binary concepts were standardized to lie between 0 and 1. 

\paragraph{Deriving Context-Aware Concepts from Unstructured Clinical Text.}
Unstructured free clinical text from discharge summaries, radiology reports and echocardiogram studies in MIMIC-IV were used to derive a set of \emph{context-aware} concepts. We use discharge summaries as opposed to admission notes because ARDS typically develops during ICU stay rather than at admission, admission notes often lack diagnostic clarity, and discharge summaries provide a complete view of the clinical trajectory and final diagnoses making them more suitable for retrospective phenotyping. Since the clinical text is authored independently of the labeling process of the patients (their diagnosis), we hypothesize this contextual information may be less susceptible to leakage and reduces the model’s reliance on structured variables (shortcuts) that may encode spurious correlations with the label. The concepts we prompt the LLM to extract are designed to capture information that is missing from the structured EHR data. Specifically, from the discharge summaries we extract mentions of physiological events and co-morbidites such as pneumonia, aspiration, pancreatitis, cardiac arrest, and transfusion-related acute lung injury (TRALI). We also ask the LLM to give an overall clinical impression of whether the patient had ARDS based on the textual information alone. In radiology reports, the LLM focused on identifying bilateral infiltrates, while in echocardiogram studies, the LLM identified cases of cardiac failure. LLM prompting details are provided in the next section.

\section{Methods}
Our goal is to train an interpretable model to classify ARDS by incorporating insights from free-text clinical notes. First, we extract structured EHR data from MIMIC-IV and define features and an initial vanilla concept set that excludes any contextual information derived from clinical notes. We then train an LLM to generate additional concept labels from clinical notes, focusing on previously unused concepts. We then augment the vanilla concept set used with these new LLM concepts into the CBM for context-awareness and use the augmented concept set for downstream prediction. We assess how augmenting the concept set affects prediction accuracy by reducing the CBM's reliance on spurious information. Finally, we discuss how erroneous concepts might be intervened upon at test time and investigate the effect of these interventions on prediction accuracy. 

\paragraph{Extracting Context-Aware LLM Concepts From Free-Text.} We obtain a set of LLM concepts ${c_{text}}$ from clinical notes as follows. We implement an LLM, configured with the Llama-3 model (trained on 7 billion parameters), using a chunk size of 4096 tokens and an overlap of 100 tokens between chunks. The prompts we use are based on earlier investigations but simplified to produce only Yes or No answers, without requesting reasoning behind the responses. After processing the responses, if any contain a ``Yes," the concept label is set to 1; otherwise, it remains 0. Notably, the LLM concepts are generated from the distribution $ p(c_{\text{text}}|x_{\text{text}})$ and are less likely to be conditionally dependent on the outcome $y$ given the input, thus reducing the risk of leakage. 

This work aims to assess whether LLM-generated concepts directly contribute to performance gains. To isolate their effect, we employ a minimal prompt template. Prompt design is known to substantially influence both concept quality and downstream model performance; therefore, using a more complex prompt would confound the source of improvement—whether due to the concepts themselves or the prompt engineering. By fixing the prompt, we control for this source of variability. Moreover, given the high computational cost of generating concepts with LLaMA, a simple prompt also reduces inference time.

We apply the prompt across different patient conditions for clarity and reusability, with minor modifications based on the specific label. For conditions such as aspiration, pneumonia, pancreatitis, cardiac arrest, and TRALI, we use the discharge summary as input and tailor the query to the condition being evaluated. We also use the summaries to get an overall clinical impression of ARDS. For bilateral infiltrates, we modify the query by replacing ``suggest" with ``mention" and use the radiology reports as input. For cardiac failure, we first assess the echocardiogram studies, and if no failure is diagnosed, we check the discharge summary for any mention of cardiac failure. The prompt format is as follows:

\begin{lstlisting}[breaklines=true]
    template=(
        Context: You are a clinician receiving chunks of clinical text for patients in an ICU. Please do the reviewing as quickly as possible.
        Task: Determine if the patient had pneumonia.
        Instructions: Answer with `Yes' or `No'. If there is not enough information, answer `No'.
        Discharge Text:{discharge_text}
        Query: Does the chunk of text suggest that the patient has pneumonia? Answer strictly in `Yes' or `No'.
    ),
    input_variables=[discharge_text]
\end{lstlisting}

\noindent The LLM concepts ${c_\text{text}}$ are then passed directly as an input to the label predictor. 

\paragraph{Problem Setup.} Let ${(x^{(i)}, y^{(i)}, c_{j}^{(i)})}_{i=1}^n$ denote a training set of input features $x$, target labels $y$ and vanilla concepts $c$. Here, input $x \in \mathbb{R}^d$ with $d=21$, consists of 15 continuous features normalized between 0-1 and 6 binary features. Target label $y \in [0,1]$ are the true classification labels corresponding to ARDS or no ARDS. As our concept set consists of both binary and continuous vanilla concepts, $c_{j}$, we define the concepts as follows. For \( j \in \{1, 2, \dots, 12\} \), \( c_j \in \mathbb{R} \) as the concepts \( c_j \) are continuous. For \( j \in \{13, 14\} \), \( c_j \in \{0, 1\} \) as the concepts \( c_j \) are binary. These concepts are derived from the tabular EHR data as discussed in Section \ref{sec:cohort}. The detailed description of the input features and concepts is present in Appendix~\ref{sec:dataset}.

Let $f(g(x))$ denote a CBM, where $g$ maps an input $x$ into the concept space and $f$ maps concepts into a final prediction. Let $L_Y$ be the loss function that measures the difference between the predicted and true target label $y$ values. Let $L_{C_j}$ be a loss function that measures the difference between the predicted and true $j$-th concept (\cite{koh2020concept}).  Finally, let the trained CBM $f(g(x))$ be represented using $\hat{f}$ and $\hat{g}$. A joint CBM trains $\hat{f}$ and $\hat{g}$ simultaneously and optimises a weighted sum of the losses for predicting concepts and the target label. That is, for a vanilla CBM:

\begin{equation}
    \hat{g},\hat{f} = \arg\min_{f,g} \sum_i \Bigl[L_Y(f(g(x^{(i)})); y^{(i)}) + \sum_j \lambda L_{C_j}(g(x^{(i)}); c^{(i)}) \Bigr]
\end{equation}

\paragraph{Context-Aware CBM Training.} For our Context-Aware CBM, we augment the predicted concepts $\hat{c}$ with the LLM-predicted concepts $c_\texttt{text}$ as this gives us additional information that the CBM can use to predict $y$ that is not artificially linked to the label $y$ as it comes from a different distribution. Our initial predicted concepts come from $p(c|x)$ whereas our LLM concepts come from a different distribution $p(c_\texttt{text}|x_\texttt{text}$). The loss function for the Context-Aware CBM thus becomes:
\begin{align}
    \hat{g},\hat{f} = \arg\min_{f,g} \sum_i \Bigl[L_Y(f(g(x^{(i)}),c_\text{text}); y^{(i)}) + \sum_j \lambda L_{C_j}(g(x^{(i)}); c^{(i)})\Bigr];   
\end{align}
% \begin{align}
%     \hat{g}, \hat{f} = \arg\min_{f, g} \sum_i \Big[ & L_Y(f(g(x^{(i)}), c_{\texttt{text}}); y^{(i)})  + \sum_j \lambda L_{C_j}(g(x^{(i)}); c^{(i)}) \Big];  
% \end{align}

\noindent Since we incorporate contextual information directly into the label loss of the context-aware CBM, this information serves as a regularization penalty for the concept representation that we learn based on the vanilla concepts. Doing so explicitly constrains the model to prevent it from overfitting to the training data distribution and reduces overreliance on potentially leaky concept information. We use one neural network to predict both intermediate concepts $c$ and final labels $y$. The network takes two inputs: vanilla features, $x$, and LLM-predicted concepts, $c_\texttt{text}$. The first part of the network $g$ predicts $\hat{c}$ from $x$ (concept predictor in Fig \ref{fig:pipeline}). This is concatenated with $c_\texttt{text}$ and put through the second part of the network $f$ to predict the label $\hat{y}$ (label predictor in Fig \ref{fig:pipeline}). The concept $L_{C_{j}}$ loss is binary cross entropy (BCE) for binary concepts and mean squared error (MSE) loss for the continuous ones.

\section{Interventions in Context-Aware CBMs}
\label{sec:interventions}
CBMs are valued for their interpretability and their support for interventions on misclassified concepts, which can improve downstream label predictions. \cite{koh2020concept} show that CBMs enable test-time interventions, allowing users to modify predicted concepts \(\hat{c}\), which in turn updates the target label prediction \(\hat{y}\). We consider three types of concept-level interventions: (1) Ground truth, where the true concept value is known (ideal but rarely feasible); (2) Mean-based, where the concept is set to its mean; and (3) Median-based, using the concept’s median. We primarily explore two strategies for intervening on these concepts.

\paragraph{Interventions assuming concepts are independent.} In this scenario, we assume the concept being intervened on does not affect the remaining concepts \cite{koh2020concept}. This is the most widely used technique, favored for its simplicity and ease of implementation. Mathematically, this is denoted as follows: let $c=[c^1,c^2,...,c^k,..,c^d]$ be the concept representation before intervention. We are intervening on the $k^{th}$ concept using $c^k_{int} = c^k_{true}/c^k_{mean}/c^k_{median}$. Here, $c^k_{true}$ is the ground-truth value, $c^k_{mean}$ is the mean value across train examples, while $c^k_{median}$ is the median value across train examples for the concept $k$. The resulting concept $c_{int}$ after intervention becomes $c_{int}=[c^1,c^2,...,c^k_{int},..,c^d]$. This naturally extends across multiple concepts, where a practitioner may choose to intervene on more than 1 concept. In such cases, the concept becomes $c_{int}=[c^1,c^2,...,c^{k_1}_{int},...,c^{k_2}_{int},..,c^d]$, where $k_1$ and $k_2$ are the indices of the concepts being intervened upon.

\paragraph{Interventions assuming correlated concepts.} This is more practical, as many concepts are correlated, and assuming independence during intervention can introduce errors: intervening on one concept may require adjustments to others. \cite{vandenhirtz2024stochasticconceptbottleneckmodels} address this by modeling a joint concept distribution, but this remains a challenging open problem. Here, we propose two techniques based on Pearson correlation between concepts:

\begin{enumerate}
    \item Correlated interventions based on one main intervention: In this scenario, we have one main concept $c^k$ being actively intervened upon, using ground-truth/mean/median values. Every other concept is intervened as follows: $c^j = c^j + \text{Corr}(c^j,c^k)\times\delta_k$, where $\delta_k = c^k_{int}-c^k$. Here, $\delta_k = c^k_{int}-c^k$ determines the magnitude of the change of concept, while $\text{Corr}(c^j,c^k)$ determines the scale of the change of concept. If $\text{Corr}(c^j,c^k) = 0$, this simplifies to the independent concept intervention. 
    \item Correlated interventions based on multiple main concept intervention: In this scenario, we have multiple main concepts $c^{k_1},c^{k_2},..,c^{k_q}$ being actively intervened upon. The other concepts are intervened as: $c^j = c^j + \text{Corr}(c^j,c^{k_1})\times\delta_{k_1} + \text{Corr}(c^j,c^{k_2})\times\delta_{k_2} + \dots + \text{Corr}(c^j,c^{k_q})\times\delta_{k_q}$ , where $\delta_{k_q} = c^{k_q}_{int}-c^{k_q}$.
\end{enumerate}
These intervention strategies are not exhaustive and can be adapted based on the use case and practitioner expertise. Our experiments illustrate how context-aware CBM performance varies with the choice of intervention. 

\section{Experimental Setup}

1. \textbf{Predictive Performance.} We evaluate the predictive performance of both the vanilla CBM and context-aware CBM on our target cohort. These models are compared against standard baseline approaches. In addition, we assess the impact of augmenting baseline models with large language model (LLM)-derived concepts to determine whether incorporating contextual information improves predictive accuracy. 2. \textbf{Reducing Leakage.} We investigate whether the context-aware CBM reduces label leakage by mitigating reliance on spurious correlations or shortcut features. To this end, we analyze mutual information and compare it across models. 3. \textbf{Interventions Analysis.} We intervene on concepts for incorrectly classified patients using techniques described in Section \ref{sec:interventions}. 4. \textbf{Unseen Data.} We test our CBMs on a cohort of data with a different distribution of patients to test the robustness of the models to distribution shift. 

\section{Evaluation Metrics}
We evaluate context-aware CBM performance against vanilla CBMs using accuracy, precision, recall, and F1-score. To assess concept quality, we use mean squared error (MSE) and mean absolute error (MAE) for continuous concepts, and accuracy and recall for binary concepts. Additionally, we compare mutual information (MI) scores to
measure how much information the concepts contribute to the final prediction. 

Formally, the MI is defined as:
\begin{align}
MI(y_{true}; y|c,x) = \sum_{y \in y_{true}} \sum_{\hat{y} \in y|c,x} P(y, \hat{y}) \log \left( \frac{P(y, \hat{y})}{P(y) P(\hat{y})} \right)
\end{align}

\noindent Here, \( P(y, \hat{y}) \) is the joint probability of the true label being \( y \) and the predicted label being \( \hat{y} \), while \( P(y) \) and \( P(\hat{y}) \) are the marginal probabilities of the true and predicted labels, respectively. A MI score of 1 indicates that the concepts capture all relevant information for predicting the final label, whereas a score of 0 suggests that the concepts do not provide any useful information. A lower score also raises the possibility of leakage, where extra input features might bypass the concept layer, undermining the model’s interpretability. Mutual information being the evaluation metric for leakage reduction has been theoretically studied in \cite{havasi2022addressing} and \cite{sun2024eliminating}. We point the reader to these references for a more detailed insight into leakage reductions for CBMs. We have also provided additional leakage metrics based on recent work (\cite{parisini2025leakage}) that considers measuring the leakage in terms of Concept Task Leakage (CTL) and Interconcept leakage (ICL), see Appendix~\ref{sec:leakage-metrics}. We evaluate the effectiveness of the interventions by checking how many false positive and false negative label predictions are corrected following the intervention.

\section{Results and Discussion} 
\vspace{-10pt}

\begin{figure}[h]
    \centering
    \includegraphics[width=0.8\linewidth] {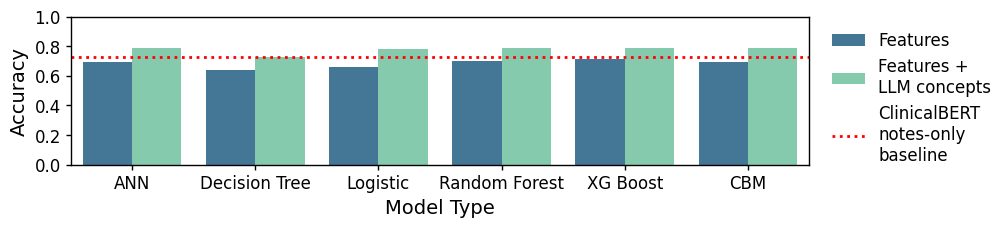}
    \caption{Comparative performance of feature-only models vs. models with LLM-derived concepts. Adding LLM concepts improves all reported metrics by $\sim$10\%. Details on ClinicalBERT baseline in Appendix \ref{sec:clinical-bert-baseline}. CBMs are competitive with baselines.}
    \label{fig:feature_to_label_llm}
\end{figure}

\begin{figure}[h]
    \centering
    \includegraphics[width=0.8\linewidth]{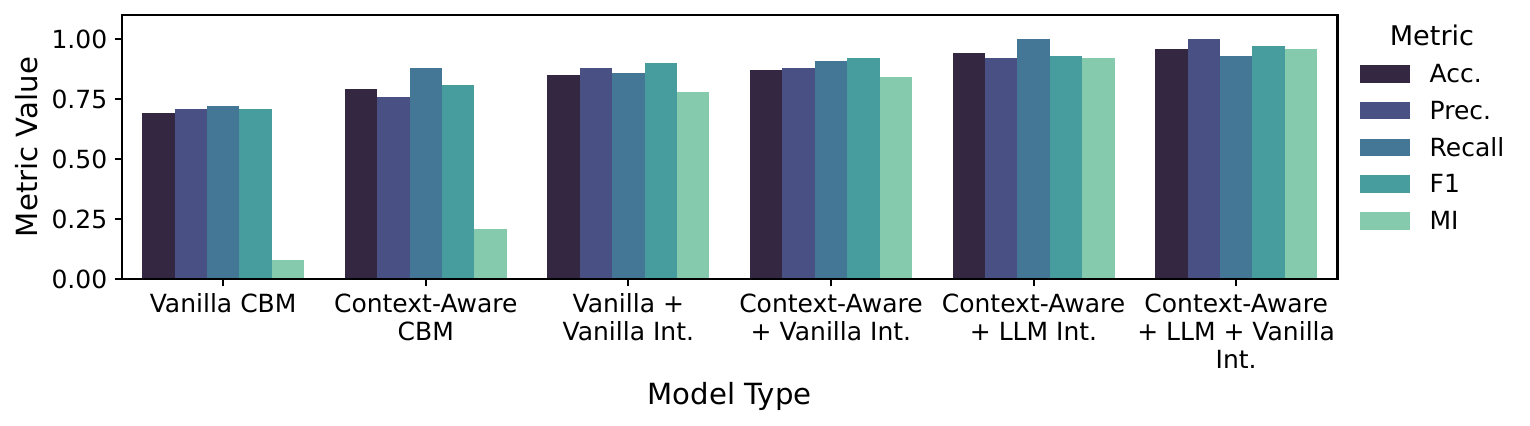}
    \caption{Performance of Vanilla and Context-Aware CBMs  after concept-level interventions on misclassified patients. Intervening on the concepts improves metrics by 12--20\%, with high gains in mutual information score.}
    \label{fig:intervention_model_performance}
\end{figure}

\begin{figure}[h]
    \centering
    \includegraphics[width=\linewidth,height=5cm]{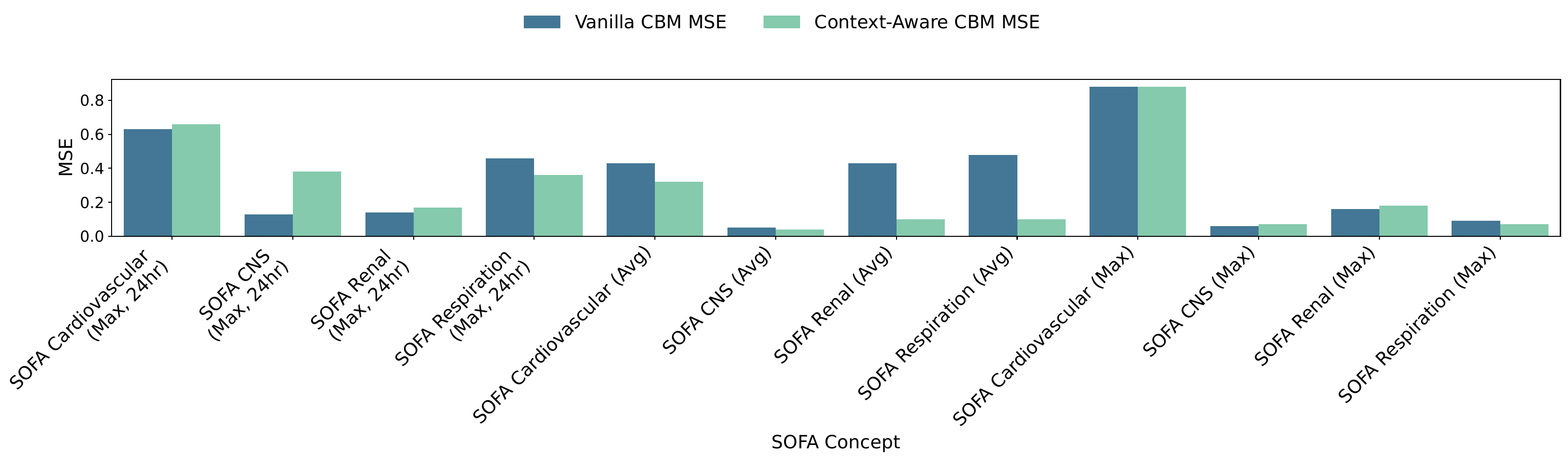}
    \caption{Comparison of concept prediction between Vanilla and Context-Aware CBMs. Context-Aware CBMs achieve lower MSE on respiration-related concepts, indicating better focus on prediction-relevant concepts and reduced leakage.}
    \label{fig:cbm_concepts}
\end{figure}

\noindent\textbf{Context-aware models augmented with LLM concepts show an 8–10\% improvement over traditional models across all metrics.} As baselines, we consider two variations of our model architecture: the first uses only physiological signals, while the second incorporates both physiological signals and LLM-extracted features as input. The results are summarized in Fig \ref{fig:feature_to_label_llm} and Table \ref{tab:feature_to_label_llm}. We observe the inclusion of LLM concepts leads to an 8–10\% improvement in ARDS labeling performance across all metrics. However, these baseline models directly map input features to final labels and do not offer insights into the reasoning behind the predictions. Thus, we extend the experiment to CBMs, which serve as the primary model architecture in subsequent experiments. From Table~\ref{tab:feature_to_label_llm}, we observe that vanilla CBMs perform comparably to all baselines while offering interpretability.

\noindent\textbf{Context-aware CBMs outperform vanilla CBMs by 10–20\% across all metrics, indicating that LLMs extract important concepts relevant for final ARDS prediction.} From Fig \ref{fig:intervention_model_performance}, we observe that context-aware CBMs augmented with LLM-derived concepts improve final ARDS prediction metrics by 10–20\%. Furthermore, Table \ref{tab:cbm_concepts} shows gains in predicting SOFA respiration scores and respiratory comorbidities, indicating stronger alignment with clinically relevant features. LLM concepts such as aspiration, bilateral infiltrate, pancreatitis, and the ARDS label are positively correlated with ARDS outcomes, while concepts like cardiac arrests and failures show negative correlations. These additions boost performance over vanilla CBMs. We analyze misclassifications by the vanilla CBM and find that context-aware CBMs correct 78\% of false negatives (FN) and 29\% of false positives (FP) by leveraging context from clinical notes. However, they also introduce 21\% new FP that were previously correct. This reflects a more conservative approach that prioritizes sensitivity, reducing missed ARDS cases at the cost of increased FP. This trade-off is acceptable, as FP can be addressed through follow-up, whereas false negatives risk delayed treatment. Performance on vanilla concepts is also detailed in Table \ref{tab:cbm_concepts}.

\begin{wraptable}{l}{0.65\linewidth}
    \centering
    \caption{Interventions across multiple concepts. We observe a higher number of corrections when intervening across correlated concepts as compared to intervening independently. Bold indicates better, with higher number of FN/FP corrections per row.}
    \label{tab:interventionscomb}
    \vspace{0.5em} % Small spacing above the table
    \footnotesize
    
    \begin{tabular}{lcccc}
        \hline
        \multirow{2}{*}{Model} & \multicolumn{2}{c}{Independent} & \multicolumn{2}{c}{Correlated} \\
        \cline{2-5}
         & FN & FP & FN & FP \\
        \hline
        Vanilla CBM: GT & 24/77 & 40/46 & \textbf{38/77} & \textbf{42/46} \\
        Vanilla CBM: Mean & 26/77 & 40/46 & \textbf{26/77} & \textbf{42/46} \\
        Vanilla CBM: Median & 24/77 & 40/46 & \textbf{30/77} & \textbf{43/46} \\
        Context-aware CBM: GT & 18/22 & 45/55 & \textbf{22/22} & \textbf{55/55} \\
        Context-aware CBM: Mean & \textbf{17/22} & 10/55 & \textbf{17/22} & \textbf{31/55} \\
        Context-aware CBM: Median & \textbf{17/22} & 10/55 & \textbf{17/22} & \textbf{35/55} \\
        \hline
    \end{tabular}\end{wraptable}

\noindent\textbf{Context-aware CBMs improve the prediction of concepts critical to the final outcome, thereby reducing leakage.} Figure~\ref{fig:cbm_concepts} and Table~\ref{tab:cbm_concepts} present statistics on the quality of intermediate concept predictions. We observe that context-aware CBMs achieve lower error on respiratory-related concepts—both continuous (based on SOFA scores) and binary (respiratory comorbidity)—compared to vanilla CBMs. These concepts are crucial for the final ARDS prediction. This suggests that augmenting with LLM-generated concepts helps assign greater emphasis to relevant features, effectively completing the concept space and mitigating information leakage.
\\\\
\noindent\textbf{Context-aware CBMs produce concepts which have higher mutual information, indicating leakage reduction.} From Fig \ref{fig:intervention_model_performance}, we observe the mutual information of context-aware CBM almost triples, indicating that LLM generated concepts add to the completeness of the concept set and reduce leakage.\\

\begin{figure}[h]
    \centering
    \includegraphics[scale=0.31]{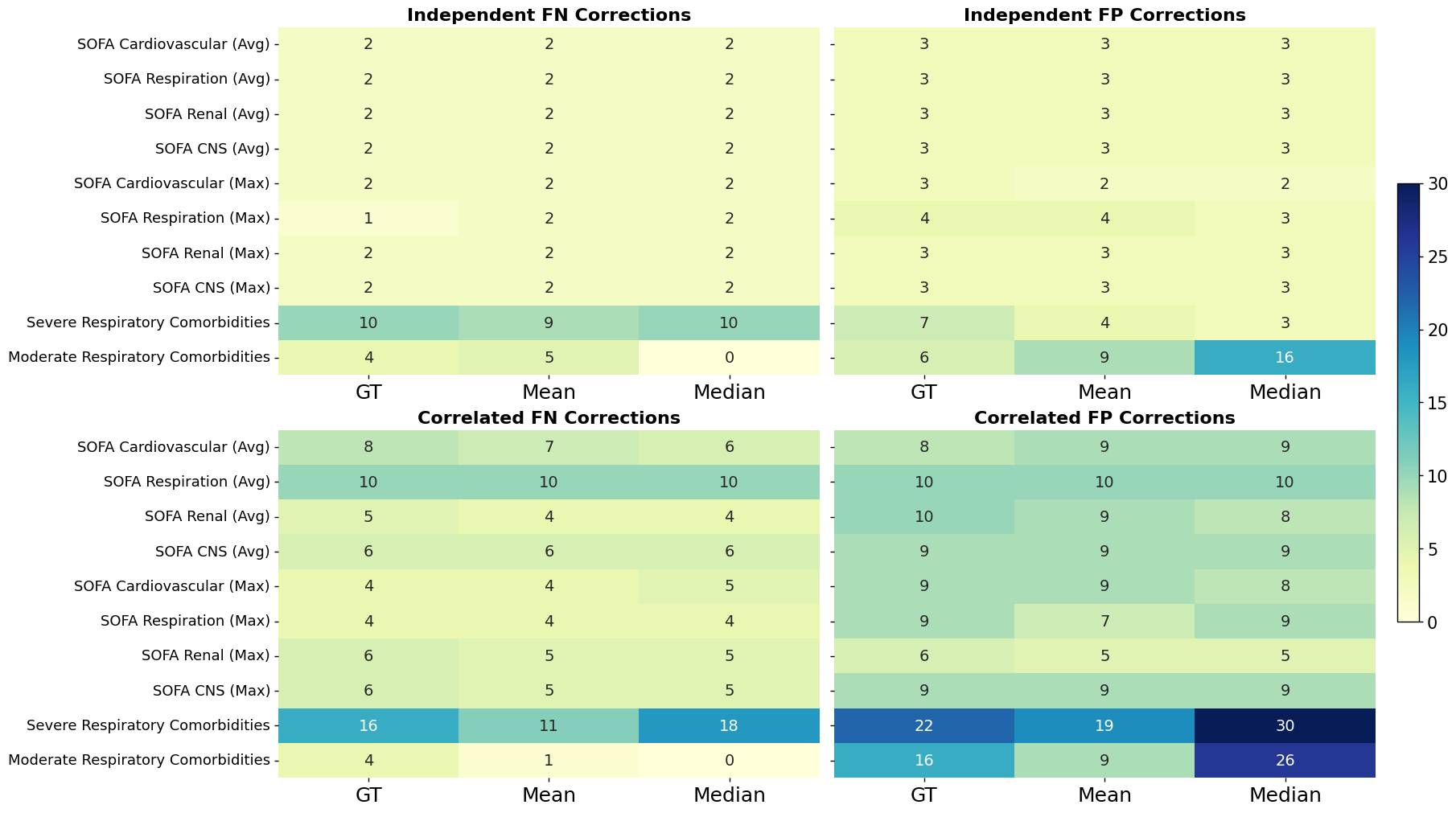}
    \caption{Interventions across individual vanilla concepts. See Figure \ref{fig:fp_fn_concepts} for LLM concepts. We observe higher number of corrections across correlated interventions as compared to independent interventions. Additionally, median interventions make higher corrections than mean interventions, indicating not all interventions are equally effective and the choice lies on the practitioner.}
    \label{fig:fp_fn_vanilla_concepts}
\end{figure}

\noindent\textbf{Interventions on misclassified concepts further boost the performance of context-aware CBMs by 12–20\%.} One of the key strengths of CBMs lies in their interpretability through concepts and the ability to intervene on incorrect concept predictions to improve overall model performance \citep{koh2020concept}. We evaluate the impact of such interventions in Fig \ref{fig:intervention_model_performance}, where we observe that intervening on context-aware CBMs results in a performance improvement of 12–20\% across all metrics. As a reminder to our readers, we focus on three types of concept-level interventions: a. Ground truth interventions, where the true value of the concept is known. b. Mean-based interventions. c. Median-based interventions. From Tables~\ref{tab:interventionscomb} and Figures \ref{fig:fp_fn_vanilla_concepts} \& \ref{fig:fp_fn_concepts}, we observe that median-based interventions outperform mean-based ones, correcting more both false positive and false negative cases. This is expected, as the median is more robust to outliers and missing data which are common challenges in medical datasets, whereas the mean is more sensitive to these issues.
\\\\
\noindent\textbf{Interventions based on correlated concepts are more effective than interventions on individual concepts.} In this section, we explore how intervention performance changes when accounting for concept correlations, as opposed to treating concepts independently. While performing multiple interventions, we select the top 3 concepts which had the highest number of corrections from the independent interventions. From Figures \ref{fig:fp_fn_vanilla_concepts} \& \ref{fig:fp_fn_concepts}, we observe that incorporating correlation awareness leads to substantial improvements, with a significantly higher number of corrections across both false negatives and false positives. Figure \ref{fig:correlations} illustrates the correlations among concepts, revealing strong relationships among various cardiovascular, respiratory, renal, and CNS-related concepts. Furthermore, as shown in Table~\ref{tab:interventionscomb}, interventions based on correlated concepts yield greater performance gains than independent interventions for both vanilla and context-aware CBMs.
\\\\
\noindent\textbf{Context-aware CBMs outperform vanilla CBMs in out-of-distribution patients.} We evaluate model performance on an unseen test cohort, as detailed in Table \ref{tab:cohort-stats}. This cohort shows notable demographic shifts, including a 10\% decrease in white patients and a 3–4\% increase in Black, African-American, Hispanic, Latino, and other racial groups. Additionally, the ARDS class distribution differs from that of the original cohort. As shown in Table \ref{tab:unseen}, context-aware CBMs outperform vanilla CBMs across all metrics, with a 38\% improvement in recall. Moreover, their performance remains consistent on the unseen cohort, showing no significant degradation compared to in-distribution test examples. This suggests that context-aware CBMs are robust to distributional shifts.
\\\\

\noindent\textbf{Discussion on performance boosts.} One might assume that adding clinical notes naturally improves performance by simply increasing data volume, but integrating multimodal data is non-trivial \citep{liu2022machine}. Clinical data is heterogeneous: vitals are structured and numerical, whereas notes are unstructured. Simple merging can introduce noise, irrelevant or missing data, and spurious correlations or leakage. We address this by using a LLM to extract key concepts, grounding inputs in clinically meaningful signals. This improves interpretability, enhances robustness, and enables targeted interventions—benefits often missing from standard multi-modal approaches \citep{liu2022multimodal}. Aligning structured and unstructured data remains an open problem \citep{islam2025llm}, one that requires more than simple concatenation and instead calls for modeling their complementary strengths. A recent ARDS model by \cite{levy2025development} used cTAKES for note processing but observed improvements primarily from radiology reports.

\begin{wraptable}{l}{0.45\linewidth}
    \centering
    \begin{tabular}{lcc}
    \toprule
    \textbf{Metric}  & \textbf{Context-aware} & \textbf{Vanilla} \\
    \midrule
    Acc.   & \textbf{0.80} & 0.68 \\
    AUC        & \textbf{0.82} & 0.66 \\
    Prec.  & \textbf{0.69} & 0.63 \\
    Recall     & \textbf{0.95} & 0.57 \\
    F1   & \textbf{0.80} & 0.60 \\
    \bottomrule
  \end{tabular}
    \caption{Performance metrics over unseen test distributions. Context-aware CBMs generalize better to out-of-distribution data than Vanilla CBMs.}
  \label{tab:unseen}
\end{wraptable}

\section{Case Studies}

We illustrate the interpretability of context-aware CBMs using case studies from our patient cohort. The models expose the key factors driving each diagnosis and let us reason about how changes to specific concepts would alter the predicted diagnosis.
\\\\
\noindent\textbf{Case study 1: Correction of false negative ARDS classifications by context-aware LLMs.} We analyze two patients incorrectly classified as non-ARDS by the vanilla CBM. Patient 1: high vasopressor requirement, low urine output, high average GCS, and no known respiratory diseases or infections. Intermediate concepts show high cardiovascular, renal, respiratory SOFA scores but no severe respiratory comorbidity, leading to a false negative. However, the LLM identifies bilateral infiltrates and pneumonia with no cardiac failure—details absent from physiological signals—correcting the classification to ARDS. Patient 2: average mean BP, high CNS GCS, and lung disease due to external factors. Intermediate concepts indicate high renal and cardiovascular SOFA scores with severe comorbidity. The LLM extracts evidence of aspiration, bilateral infiltrates, and pneumonia with no cardiac failure from clinical notes, enabling correct ARDS classification.
\\\\
\noindent\textbf{Case study 2: Correction of false positive ARDS classifications by context-aware LLMs.} We examine two patients incorrectly classified as ARDS-positive by the vanilla CBM. Patient 1: low vasopressor requirement, moderate urine output, high creatinine, high average GCS, with influenza and chronic lower respiratory disease. Intermediate concepts show high cardiovascular, respiratory, CNS SOFA scores, moderate respiratory comorbidity, and high renal SOFA score in the first 24 hours—leading to a false positive. The LLM identifies cardiac failure, not reflected in physiological signals, correcting the diagnosis to non-ARDS. Patient 2: average mean BP, low urine output, high creatinine, high GCS, with influenza. Predicted concepts include high respiratory, renal, cardiovascular, and CNS SOFA scores with severe comorbidity. The LLM detects cardiac arrest and failure from clinical notes, correcting the classification to negative ARDS.
\\\\
\noindent\textbf{Case study 3: LLM concepts can induce misclassifications, correctable via concept-level interventions.} We analyze two patients where Context-Aware CBMs introduced errors absent in the vanilla CBM predictions. Patient 1: moderate vasopressor requirement and BP, high urine output, low creatinine, high average GCS. Predicted concepts include low cardiovascular and renal SOFA scores, high respiratory SOFA score, and moderate respiratory comorbidity. The vanilla CBM correctly classifies the patient as non-ARDS. However, the LLM detects pneumonia and ARDS in clinical notes, inducing a false positive. Intervening on the ARDS detected concept restores the correct non-ARDS classification. Patient 2: low vasopressor requirement, normal BP, low urine output, high creatinine, with influenza and other respiratory diseases. Predicted concepts include high respiratory, renal, and CNS SOFA scores with severe comorbidity. The vanilla CBM correctly classifies the patient as ARDS-positive. However, the LLM introduces a false cardiac arrest concept, leading to a false negative. Correcting this concept restores the correct ARDS-positive prediction. Hence, we show that even if LLMs introduce leakage, our method makes it easy to fix by leveraging CBMs’ ability to trace and edit individual concepts.
\\\\
\noindent\textbf{Case study 4: Concept-based interpretability enables correction of CBM misclassifications.} We examine two patients misclassified by both vanilla and Context-Aware CBMs, highlighting how interpreting intermediate concepts helps identify and correct errors. First, a true positive patient: low vasopressor requirement, normal BP, moderately impaired PaO2-FiO2 ratio, very high urine output, high GCS, with chronic lower respiratory and external agent-induced lung diseases. Predicted correctly: respiratory and CNS SOFA scores; incorrectly: renal SOFA and severe morbidity. LLM identifies aspiration, bilateral infiltrates, and pneumonia. Correcting the morbidity concept aligns both CBMs with the true ARDS-positive label. Second, a true negative patient: low vasopressor requirement, normal BP, moderately impaired PaO2-FiO2 ratio, high urine output, creatinine, and GCS, with other respiratory diseases. Predicted correctly: cardiovascular, respiratory, and renal SOFA scores; incorrectly: CNS SOFA and morbidity. LLM fails to capture cardiac failure. The combined effect of incorrect morbidity and missing cardiac failure leads to misclassification. Intervening on these concepts corrects both CBMs to the true negative label.

\section{Conclusions and Limitations}
We developed a context-aware concept bottleneck model (CBM) that integrates structured EHR data with unstructured clinical notes processed by a large language model (LLM). This hybrid approach improves both the accuracy and interpretability of retrospective ARDS classification by surfacing clinically meaningful concepts often missed by traditional signals. The CBM’s transparency enables physician intervention to correct errors, reducing false positives/negatives, mitigating leakage and shortcuts, improving concept completeness, and strengthening the diagnostic pipeline. Future work includes training models on time-limited data to reflect real-world diagnostic constraints, as our study benefited from full patient timelines. Additionally, while LLM-derived concepts provide valuable signals, they may introduce noise or hallucinations. Hence, clinician oversight will be essential for validation prior to deployment.

\newpage

\bibliographystyle{plainnat}
\bibliography{literature}

\newpage

\section*{Acknowledgements}

We thank Shreya Kamath for designing Figure \ref{fig:pipeline}. Your contribution helped define the face of this paper and your fresh perspective was very insightful.

\appendix

\section{Cohort Selection and Preprocessing}
\label{sec:dataset}

\noindent\textbf{Feature Selection}

\begin{itemize}
    \item \textbf{Features from SOFA Scores and Related Lab Measurements:} SOFA scores (Sequential Organ Failure Assessment) are useful in ARDS diagnosis because ARDS is a systemic condition, not just a respiratory issue—it often arises as part of multi-organ dysfunction, typically in critically ill patients. Of the six SOFA components, we selected four considered most relevant for ARDS characterisation: respiratory, cardiovascular, renal, and CNS. The features were extracted from the MIMIC-derived SOFA tables. For each organ component, we included the lab measurements used to calculate the corresponding SOFA score (e.g. creatinine and urine output for renal). Specifically, we extracted the lab values associated with the worst hourly SOFA score across the stay, the average SOFA score across the stay, and the worst hourly score within the first 24 hours. All of these were treated as continuous features.
    \item \textbf{Features from Norepinephrine:} Norepinephrine equivalent dose was included as a feature because it provides a single, standardised measure of the total vasopressor support a patient required. To capture both overall exposure and peak intensity, we calculated the time-weighted average norepinephrine equivalent dose across the stay, as well as the maximum dose administered. These were continuous features.
    \item \textbf{Feature from Mechanical Ventilation Duration:} Another feature included was the total duration the patient spent on mechanical ventilation during their hospital stay, as this serves as a direct indicator of respiratory support needs and is clinically relevant to ARDS severity. This was a continuous feature.
    \item \textbf{Features from Pre-existing Respiratory Conditions:} We also identified whether patients had any pre-existing respiratory conditions by extracting ICD-9 and ICD-10 diagnosis codes from their hospital admissions. These diagnoses were grouped into six respiratory illness categories: upper respiratory infections, influenza and pneumonia, acute lower respiratory infections, chronic lower respiratory diseases, lung diseases due to external agents, and other respiratory diseases. The features were binary.
    \item \textbf{Feature Pre-processing:} We used a mixture of continuous and binary features. We removed 6 features that were missing values for over 50\% of patients (5 first 24hr measurements and 1 creatinine measurement, highlighted in Section \ref{sec:dataset}). Then, any missing values were imputed using the median calculated from all patients in the training set. We chose to use the median as it is less sensitive to outliers and there are many patients with extreme values on either end of the ranges for several features. All continuous features were then scaled using min-max scaling to be between 0 and 1. 
\end{itemize}

\noindent Some features had to be removed as their values were missing for more than 50\% of patients.\\

\noindent SOFA Scores - Continuous Values
\begin{enumerate}
    \item Worst GCS value for CNS
    \item Worst first 24hr GCS value for CNS (removed)
    \item Average GCS value for CNS
    \item Worst creatinine value for Renal (removed)
    \item Worst urine output value for Renal
    \item Worst first 24hr creatinine value for Renal (removed)
    \item Worst first 24hr urine output value for Renal (removed)
    \item Average creatinine value for Renal
    \item Average urine output value for Renal
    \item Worst PaO2/FiO2 ratio for Respiration
    \item Worst first 24hr PaO2/FiO2 ratio for Respiration (removed)
    \item Average PaO2/FiO2 ratio for Respiration
    \item Worst norepinephrine rate for Cardiovascular
    \item Worst mean blood pressure for Cardiovascular
    \item Worst first 24hr mean blood pressure for Cardiovascular
    \item Worst first 24hr norepinephrine rate for Cardiovascular (removed)
    \item Average norepinephrine rate for Cardiovascular
    \item Average mean blood pressure for Cardiovascular
\end{enumerate}

\noindent Norepinephrine Equivalent Dosage - Continuous Values
\begin{enumerate}
    \item Average Norepinephrine Dosage
    \item Peak Norepinephrine Dosage
\end{enumerate}

\noindent Mechanical Ventilation Duration - Continuous Value
\begin{enumerate}
    \item Minutes Patient was on Mechanical Ventilation
\end{enumerate} 

\noindent Pre-existing Conditions - Binary Values
\begin{enumerate}
    \item Upper Respiratory Infections – ICD-9: 460–466, ICD-10: J00–J06
    \item Influenza and Pneumonia – ICD-9: 480–487, ICD-10: J09–J18
    \item Acute Lower Respiratory Infections – ICD-9: 466–469, ICD-10: J20–J22
    \item Chronic Lower Respiratory Diseases – ICD-9: 490–496, ICD-10: J40–J47
    \item Lung Diseases Due to External Agents – ICD-9: 500–508, ICD-10: J60–J70
    \item Other Respiratory Diseases – ICD-9: 510–519, ICD-10: J80–J99
\end{enumerate}

\noindent\textbf{Vanilla Concepts}

\noindent SOFA - Continuous Values

\begin{enumerate}
    \item Worst CNS Score across patient stay
    \item Worst CNS Score in first 24 hours
    \item Average CNS Score
    \item Worst Renal Score across patient stay
    \item Worst Renal Score in first 24 hours
    \item Average Renal Score
    \item Worst Respiration Score across patient stay
    \item Worst Respiration Score in first 24 hours
    \item Average Respiration Score
    \item Worst Cardiovascular Score across patient stay
    \item Worst Cardiovascular Score in first 24 hours
    \item Average Cardiovascular Score
\end{enumerate}

\noindent Pre-existing Conditions - Binary Values

\begin{enumerate}
    \item Moderate pre-existing respiratory comorbidity
    \item Severe pre-existing respiratory comorbidity
\end{enumerate}

\noindent\textbf{LLM Concepts}

\noindent Whether text suggests patient has each of the following conditions (all binary labels):
\begin{enumerate}
    \item Pneumonia
    \item Aspiration
    \item Pancreatitis
    \item Cardiac Arrest
    \item TRALI
    \item Overall impression of ARDS
    \item Bilateral Infiltrates
    \item Cardiac Failure
\end{enumerate}

\newpage

\section{Additional Leakage Metrics}
\label{sec:leakage-metrics}
There is recent work that considers measuring the leakage in terms of Concept Task Leakage (CTL) and Interconcept Leakage (ICL) \citep{parisini2025leakage}.  CTL measures the additional task-relevant information encoded in the predicted concepts that is not present in the ground-truth concepts. ICL measures the extra information each concept prediction carries about other concepts (cross-concept dependence). Empirically, both metrics are sensitive to leakage and strongly correlate with the effectiveness of concept-level interventions. In practice, CTL can boost task performance by allowing the model to exploit task signals embedded in the concept predictors, whereas ICL provides redundant pathways that enable high task scores even when individual concept predictions are poor \citep{heidemann2023concept}.
Concept-task leakage is defined as follows:

\[
\mathrm{CTL}_i(\hat{c}, c, y) = \max\left(0, \frac{I(\hat{c}_i, y)}{H(y)} - \frac{I(c_i, y)}{H(y)}\right)
\]
where $\mathbf{\hat{c}_{i}}$ is the predicted concepts and $\mathbf{c_{i}}$ are the ground truth concepts. This is the difference in mutual information between the predicted and ground truth mutual information (MI) between the concept $\mathbf{i}$ and the task label. The CTL score is \( 0 \leq \mathrm{CTL}_i \leq 1 \). Note that a negative difference indicates poor concept learning rather than leakage as it means that the learned concepts are less predictive of the task than we would expect from the ground truth.

Interconcept leakage is defined as the pairwise mutual information between concepts $\mathbf{i}$ and $\mathbf{j}$:

\[
\mathrm{ICL}_{ij}(\hat{c}, c) = \max\left(0,\ \frac{I(\hat{c}_i, \hat{c}_j)}{\sqrt{H(\hat{c}_i)\, H(\hat{c}_j)}} - \frac{I(c_i, c_j)}{\sqrt{H(c_i)\, H(c_j)}}\right)
\]

This metric again is \( 0 \leq \mathrm{ICL}_i \leq 1 \) and $\mathbf{ICL_{ii} = 0}$ and it quantifies the extra information that each concept encodes about the other concepts. We performed additional experiments to validate our approach in terms of CTL and ICL. CTL did not materially differ between models, indicating similar amounts of task information encoded beyond the ground-truth concepts. In contrast, ICL was lower for the context-aware CBM, indicating reduced cross-concept information sharing. Under the sufficiency rule of \cite{parisini2025leakage}, if either CTL or ICL is higher for model A than model 
B while the other metric is comparable, then A exhibits greater leakage. Our results imply the context-aware CBM has less leakage than the vanilla CBM. See Table~\ref{tab:ctl_icl_scores} for the CTL and ICL scores.

\begin{table*}[h]
\centering
\resizebox{\textwidth}{!}{%
\begin{tabular}{p{6cm}cccc}
\toprule
\textbf{Concept} & \textbf{CTL (Vanilla)} & \textbf{CTL (Context-Aware)} & \textbf{ICL (Vanilla)} & \textbf{ICL (Context-Aware)} \\
\midrule
c\_first24hr\_sofa\_max\_cardiovascular & 0 & 0.009 & 0.0231 & 0.0135 \\
c\_first24hr\_sofa\_max\_cns            & 0 & 0.0098 & 0.0319 & 0.0083 \\
c\_first24hr\_sofa\_max\_renal          & 0.0235 & 0.0006 & 0.0173 & 0.0147 \\
c\_first24hr\_sofa\_max\_respiration    & 0.0293 & 0.029 & 0.0073 & 0.0200 \\
c\_mod\_resp\_comorbidity               & 0.0102 & 0.0152 & 0 & 0 \\
c\_sofa\_avg\_cardiovascular            & 0 & 0.0427 & 0.0211 & 0.0181 \\
c\_sofa\_avg\_cns                       & 0.0063 & 0.0106 & 0.0269 & 0.0045 \\
c\_sofa\_avg\_renal                     & 0.0772 & 0.0508 & 0.0112 & 0.0086 \\
c\_sofa\_avg\_respiration               & 0 & 0 & 0.0083 & 0.0203 \\
c\_sofa\_max\_cardiovascular            & 0 & 0.0089 & 0.0104 & 0.0081 \\
c\_sofa\_max\_cns                       & 0.0213 & 0 & 0.0235 & 0.0075 \\
c\_sofa\_max\_renal                     & 0 & 0.0689 & 0.0218 & 0.0094 \\
c\_sofa\_max\_respiration               & 0 & 0.0355 & 0.0111 & 0.0040 \\
c\_svr\_resp\_comorbidity               & 0 & 0 & 0 & 0 \\
\bottomrule
\end{tabular}%
}
\caption{CTL and ICL scores for Vanilla and Context-Aware CBMS across different concepts. 0 is the lowest score possible, and lower scores are better as they indicate less leakage.}
\label{tab:ctl_icl_scores}
\end{table*}

\section{Clinical Bert Baseline}
\label{sec:clinical-bert-baseline}

We also tried extracting the additional concepts from the clinical notes using ClinicalBert instead of the LLaMa model. In order to enable a fair comparison between the two models, we tried to keep the data-preprocessing as similar as possible. For example, we kept the same decision rule: a patient is classified as having concept if any of their chunks is predicted positive. However, some changes had to be made. We set the chunk size to 512 tokens to match BERT’s maximum input length. As ClinicalBERT is not a natural language model, it could not be prompted like the LLaMa model had been. We instead computed cosine similarity scores between each chunk's embedding and two prototype sentences, one which was positive for the condition and one which was negative. For example, `This patient has pneumonia' and `This patient does not have pneumonia'. The label was assigned based on which prototype sentence the chunk was more semantically similar to. In Tables \ref{tab:agreement_rates}, \ref{tab:cbm_concepts_bert}, and \ref{tab:cbm_label_metrics_bert}, we outline and compare the ClinicalBERT results. The Context-Aware CBM using ClinicalBERT concepts yields only slightly better recall than a Vanilla CBM but performs worse on all other metrics. The Context-Aware CBM using LLaMa-derived concepts outperforms the Context-Aware CBM with BERT concepts on all metrics by 7-24\%. This is because ClinicalBERT performs poorly without labels (we do not have ground truth labels for the LLM concepts) and does not perform reasoning like an LLM does, it just produces embeddings from the text.

In Table \ref{tab:cbm_label_metrics_bert}, we show the performance of three ClinicalBERT baselines that were set up as follows. Firstly, for the features only baseline, all features were pre-processed as described in Appendix \ref{sec:dataset}. Each feature was transformed into a line like `Cardiovascular SOFA – worst mean BP 59 mmHg, avg mean BP 61 mmHg'. Each line was tokenized and passed to a pretrained ClinicalBERT model. The model was fine-tuned for binary classification using the ARDS labels in the training dataset. The ClinicalBERT in this setting underperforms compared to all other baselines as it is trained on real, natural language. The sentences generated lack the context of rich clinical notes. For the notes only baseline, each note was split into overlapping chunks. The ClinicalBERT model was fine-tuned for binary classification using the chunked inputs. This ClinicalBERT model achieves competitive accuracy but it is not interpretable as it does not provide any explanations or intermediate concepts. The final baseline uses the ClinicalBERT note embeddings for each patient and concatenates this with the raw tabular features. This was inputted into a simple two layer neural network with a sigmoid activation function for binary classification.

\begin{table}[h]
\centering
\caption{Agreement rates between ClinicalBERT and Llama for each LLM concept.}
\label{tab:agreement_rates}
\begin{tabular}{l c}
\toprule
\textbf{Concept} & \textbf{Agreement Rate (\%)} \\
\midrule
ARDS Mention        & 50.74 \\
Aspiration          & 65.95 \\
Bilateral infiltrates & 82.33 \\
Cardiac arrest      & 70.10 \\
Cardiac failure     & 59.75 \\
Pancreatitis        & 75.06 \\
Pneumonia           & 32.67 \\
TRALI               & 54.66 \\
\bottomrule
\end{tabular}
\end{table}

\begin{table*}[h]
    \centering
    \resizebox{\textwidth}{!}{%
    \begin{tabular}{p{5cm}cccc}
        \toprule
        \textbf{Concept} & \multicolumn{2}{c}{\textbf{Vanilla CBM}} & \multicolumn{2}{c}{\textbf{Context-Aware CBM}} \\
        \cmidrule(lr){2-3} \cmidrule(lr){4-5}
        & \textbf{MSE} & \textbf{MAE} & \textbf{MSE} & \textbf{MAE} \\
        \midrule 
SOFA Cardiovascular (Max, 24hr) & $0.63 \pm 0.203$ & $0.41 \pm 0.081$ & $\mathbf{0.47 \pm 0.064}$ & $\mathbf{0.36 \pm 0.028}$ \\ 
SOFA CNS (Max, 24hr)            & $\mathbf{0.13 \pm 0.032}$ & $\mathbf{0.26 \pm 0.037}$ & $0.16 \pm 0.116$ & $0.29 \pm 0.129$ \\ 
SOFA Renal (Max, 24hr)          & $\mathbf{0.14 \pm 0.065}$ & $\mathbf{0.27 \pm 0.062}$ & $0.51 \pm 0.787$ & $0.36 \pm 0.182$ \\ 
SOFA Respiration (Max, 24hr)    & $0.46 \pm 0.714$ & $0.38 \pm 0.178$ & $\mathbf{0.10 \pm 0.020}$ & $\mathbf{0.24 \pm 0.014}$ \\ 
SOFA Cardiovascular (Avg)       & $0.43 \pm 0.220$ & $\mathbf{0.24 \pm 0.059}$ & $\mathbf{0.41 \pm 0.323}$ & $0.30 \pm 0.119$ \\ 
SOFA CNS (Avg)                  & $\mathbf{0.05 \pm 0.058}$ & $\mathbf{0.12 \pm 0.089}$ & $0.28 \pm 0.499$ & $0.21 \pm 0.167$ \\ 
SOFA Renal (Avg)                & $0.43 \pm 0.689$ & $0.34 \pm 0.214$ & $\mathbf{0.08 \pm 0.022}$ & $\mathbf{0.20 \pm 0.014}$ \\ 
SOFA Respiration (Avg)          & $0.48 \pm 0.865$ & $0.31 \pm 0.243$ & $\mathbf{0.37 \pm 0.493}$ & $\mathbf{0.29 \pm 0.262}$ \\ 
SOFA Cardiovascular (Max)       & $0.88 \pm 0.526$ & $0.37 \pm 0.157$ & $\mathbf{0.62 \pm 0.140}$ & $\mathbf{0.28 \pm 0.007}$ \\ 
SOFA CNS (Max)                  & $\mathbf{0.06 \pm 0.036}$ & $\mathbf{0.14 \pm 0.036}$ & $0.07 \pm 0.060$ & $0.19 \pm 0.090$ \\ 
SOFA Renal (Max)                & $0.16 \pm 0.054$ & $0.29 \pm 0.084$ & $\mathbf{0.13 \pm 0.030}$ & $\mathbf{0.24 \pm 0.020}$ \\ 
SOFA Respiration (Max)          & $\mathbf{0.09 \pm 0.073}$ & $\mathbf{0.23 \pm 0.126}$ & $0.16 \pm 0.232$ & $0.24 \pm 0.141$ \\ 
        \bottomrule
        \textbf{Concept} & \multicolumn{2}{c}{\textbf{Vanilla CBM}} & \multicolumn{2}{c}{\textbf{Context-Aware CBM}} \\
        \cmidrule(lr){2-3} \cmidrule(lr){4-5}
        & \textbf{Accuracy} & \textbf{Recall} & \textbf{Accuracy} & \textbf{Recall} \\
        \midrule 
Respiratory Comorbidity (Moderate) & $0.85 \pm 0.048$ & $0.96 \pm 0.000$ & $\mathbf{0.89 \pm 0.035}$ & $\mathbf{0.95 \pm 0.002}$ \\ 
Respiratory Comorbidity (Severe)  & $0.98 \pm 0.002$ & $0.95 \pm 0.005$ & $\mathbf{0.98 \pm 0.002}$ & $\mathbf{0.95 \pm 0.009}$ \\
        \bottomrule        
    \end{tabular}%
    }
    \caption{Concept prediction performance for Vanilla CBM and Context-Aware CBM using ClinicalBERT concepts. The metrics used are MSE and MAE for regression tasks, and accuracy and recall for classification tasks. Bold indicates better performance.}
    \label{tab:cbm_concepts_bert}
\end{table*}

\begin{table}[h]
\centering
\begin{tabular}{lccccc}
\toprule
\textbf{Metric} & \shortstack{Vanilla\\CBM} & \shortstack{Context-Aware\\CBM} & \shortstack{BERT\\Features} & \shortstack{BERT\\Notes} & \shortstack{BERT Features\\+ Notes}\\
\midrule
Precision & $0.72$ & $0.71$ & $0.53$ & $0.72$ & $\mathbf{0.75}$ \\
Recall    & $0.69$ & $0.71$ & $\mathbf{1.00}$ & $0.86$ & $0.88$ \\
F1 Score  & $0.70$ & $0.70$ & $0.69$ & $0.79$ & $\mathbf{0.81}$ \\
Accuracy  & $0.69$ & $0.68$ & $0.53$ & $0.72$ & $\mathbf{0.78}$ \\
\bottomrule
\end{tabular}
\caption{Comparison of label prediction performance between Vanilla CBM and Context-Aware CBM using the ClinicalBERT concepts, ClinicalBERT using text generations of the raw features, ClinicalBERT using just the clinical notes, and ClinicalBERT using both clinical note embeddings and raw features concatenated and put through a simple neural network. Bold indicates better performance.}
\label{tab:cbm_label_metrics_bert}
\end{table}

\section{LLM-Derived Concepts Study}
We conducted ablation studies to examine how performance changes when using individual LLM concepts in the context-aware CBM. The results in Table \ref{tab:ablation_study} show that each concept contributes useful signals, with pneumonia and ARDS being especially impactful. Pneumonia yields the highest accuracy, followed by mention of ARDS and bilateral infiltrates. Individually, concepts achieve strong recall (e.g., 0.91 for bilateral infiltrates), but best performance is achieved when they are combined. Removal of even a single concept leads to performance drops, indicating that each captures complementary information.

To validate the LLM-derived concepts, labels for pneumonia, pancreatitis, and cardiac failure were evaluated against expert annotations (full validation was infeasible), see Table \ref{tab:llm_concept_validation}. The ARDS labels were clinician created and had 86\% agreement ($K =$ 0.76) on a sample of 100 patients. For pneumonia and cardiac failure, the LLM achieved strong F1 scores (0.71 and 0.67 respectively) and high recall ($>$0.9), supporting their utility in capturing meaningful clinical signals. Pancreatitis, though much rarer (2.71\% prevalence), had high recall (0.96) but low precision (0.24), suggesting that while the LLM may over-include marginal cases, it successfully captures most true positives.

\begin{table*}[h]
\centering
\resizebox{\textwidth}{!}{%
\begin{tabular}{p{5cm}ccccc}
\toprule
\textbf{LLM Concept} & \textbf{Accuracy} & \textbf{AUC} & \textbf{Precision} & \textbf{Recall} & \textbf{F1 Score} \\
\midrule
ARDS Mention           & 0.73 & 0.72 & 0.69 & 0.89 & 0.78 \\
Aspiration             & 0.68 & 0.67 & 0.66 & 0.82 & 0.73 \\
Bilateral Infiltrates   & 0.69 & 0.67 & 0.65 & 0.91 & 0.76 \\
Cardiac Arrest         & 0.68 & 0.66 & 0.65 & 0.84 & 0.73 \\
Cardiac Failure        & 0.70 & 0.70 & 0.71 & 0.75 & 0.73 \\
Pancreatitis           & 0.68 & 0.67 & 0.65 & 0.84 & 0.74 \\
Pneumonia              & 0.76 & 0.76 & 0.75 & 0.84 & 0.79 \\
TRALI                  & 0.67 & 0.65 & 0.64 & 0.87 & 0.74 \\
\bottomrule
\end{tabular}%
}
\caption{We ran ablation studies to assess the performance impact of individual LLM-derived concepts within the Context-Aware CBM.}
\label{tab:ablation_study}
\end{table*}

\begin{table*}[h]
\centering
\resizebox{\textwidth}{!}{%
\begin{tabular}{p{5cm}ccccc}
\toprule
\textbf{LLM Concept} & \textbf{Accuracy} & \textbf{Precision} & \textbf{Recall} & \textbf{F1 Score} & \textbf{Prevalence} \\
\midrule
Cardiac Failure        & 0.69 & 0.53 & 0.92 & 0.67 & 34.38\% \\
Pancreatitis           & 0.92 & 0.24 & 0.96 & 0.38 & 2.71\% \\
Pneumonia              & 0.69 & 0.56 & 0.97 & 0.71 & 39.80\% \\
\bottomrule
\end{tabular}%
}
\caption{Three LLM concepts were validated against expert-derived labels. }
\label{tab:llm_concept_validation}
\end{table*}

\section{Additional Case Study} 
\noindent\textbf{Case study:} We consider two patients with the following features. Patient 1: normal blood pressure, moderately impaired PaO2-FiO2 ratio, very high urine output, normal creatinine, no influenza detected, long ventilation duration, and lung disease due to external agents. Predicted intermediate concepts include high cardiovascular, respiratory, renal, and CNS SOFA scores, indicating high morbidity. Patient 2: normal blood pressure, moderately impaired PaO2-FiO2 ratio, normal urine output, very high creatinine, long ventilation duration, and pneumonia-influenza with lung disease due to external agents. Predicted intermediate concepts are similar, with high morbidity. Despite differences in patient features and non-respiratory concepts, the Context-Aware CBM correctly identifies the respiratory symptoms and classifies both patients as having ARDS.

\section{Additional Results}

\begin{itemize}
    \item Table \ref{tab:feature_to_label_llm} gives the tabulated results that correspond to Figure \ref{fig:feature_to_label_llm} but with all standard errors included. This experiment was to show that the inclusion of LLM concepts led to improvement on label prediction metrics.
    \item Table \ref{tab:cbm_concepts} provides the corresponding tabulated results for Figure \ref{fig:cbm_concepts}.
    \item Table \ref{tab:joint_vs_sequential} compares the performance of joint and sequential CBMs to confirm that leakage does not stem from the training method.
\end{itemize}

\begin{table}[h]
    \centering
    \resizebox{\textwidth}{!}{%
    \begin{tabular}{p{2.5cm}lccccc}
        \toprule
        \textbf{Model Type} & \textbf{Predictor} & \textbf{Accuracy} & \textbf{AUC} & \textbf{Precision} & \textbf{Recall} & \textbf{F1 Score} \\
        \midrule
        \multirow{2}{*}{ANN}       
            & Features                  & $0.69 \pm 0.002$ & $0.69 \pm 0.001$ & $0.71 \pm 0.003$ & $0.71 \pm 0.010$ & $0.71 \pm 0.004$ \\
            & Features + LLM concepts   & $\mathbf{0.79 \pm 0.007}$ & $\mathbf{0.79 \pm 0.007}$ & $\mathbf{0.78 \pm 0.004}$ & $\mathbf{0.84 \pm 0.011}$ & $\mathbf{0.81 \pm 0.007}$ \\
        \midrule
        \multirow{2}{*}{Decision Tree}       
            & Features                  & $0.64 \pm 0.006$ & $0.63 \pm 0.006$ & $0.64 \pm 0.006$ & $0.64 \pm 0.006$ & $0.64 \pm 0.006$ \\
            & Features + LLM concepts   & $\mathbf{0.73 \pm 0.004}$ & $\mathbf{0.73 \pm 0.004}$ & $\mathbf{0.73 \pm 0.004}$ & $\mathbf{0.73 \pm 0.004}$ & $\mathbf{0.73 \pm 0.004}$ \\
        \midrule
        \multirow{2}{*}{Logistic}       
            & Features                  & $0.66 \pm 0.011$ & $0.65 \pm 0.011$ & $0.65 \pm 0.010$ & $0.76 \pm 0.020$ & $0.70 \pm 0.011$ \\
            & Features + LLM concepts   & $\mathbf{0.78 \pm 0.009}$ & $\mathbf{0.78 \pm 0.009}$ & $\mathbf{0.74 \pm 0.010}$ & $\mathbf{0.90 \pm 0.020}$ & $\mathbf{0.82 \pm 0.008}$ \\
        \midrule
        \multirow{2}{*}{Random Forest}       
            & Features                  & $0.70 \pm 0.007$ & $0.70 \pm 0.006$ & $0.70 \pm 0.007$ & $0.70 \pm 0.007$ & $0.70 \pm 0.007$ \\
            & Features + LLM concepts   & $\mathbf{0.79 \pm 0.006}$ & $\mathbf{0.78 \pm 0.006}$ & $\mathbf{0.79 \pm 0.006}$ & $\mathbf{0.79 \pm 0.006}$ & $\mathbf{0.78 \pm 0.006}$ \\
        \midrule
        \multirow{2}{*}{XG Boost}       
            & Features                  & $0.71 \pm 0.007$ & $0.70 \pm 0.006$ & $0.71 \pm 0.009$ & $0.71 \pm 0.007$ & $0.71 \pm 0.006$ \\
            & Features + LLM concepts   & $\mathbf{0.79 \pm 0.006}$ & $\mathbf{0.79 \pm 0.006}$ & $\mathbf{0.79 \pm 0.006}$ & $\mathbf{0.79 \pm 0.006}$ & $\mathbf{0.79 \pm 0.006}$ \\
        \midrule
        Vanilla CBM        & Features         & $0.69 \pm 0.011$ & $0.69 \pm 0.009$ & $0.72 \pm 0.048$ & $0.69 \pm 0.120$ & $0.70 \pm 0.044$ \\
        Context-Aware CBM  & Features + LLM concepts &  $\mathbf{0.79 \pm 0.008}$ &  $\mathbf{0.78 \pm 0.006}$ &  $\mathbf{0.76 \pm 0.021}$ &  $\mathbf{0.89 \pm 0.058}$ &  $\mathbf{0.82 \pm 0.015}$ \\
        \bottomrule
    \end{tabular}%
    }
    \caption{Corresponding tabulated results for Figure \ref{fig:feature_to_label_llm}. Comparative performance of feature-only models vs. models with LLM-derived concepts. Adding LLM concepts improves all reported metrics by around 10\%. Additionally, CBMs perform competitively against baselines while being interpretable and intervenable. Here, bold indicates better.}
    \label{tab:feature_to_label_llm}
\end{table}

\begin{table*}[h]
    \centering
    \resizebox{\textwidth}{!}{%
    \begin{tabular}{p{5cm}cccc}
        \toprule
        \textbf{Concept} & \multicolumn{2}{c}{\textbf{Vanilla CBM}} & \multicolumn{2}{c}{\textbf{Context-Aware CBM}} \\
        \cmidrule(lr){2-3} \cmidrule(lr){4-5}
        & \textbf{MSE} & \textbf{MAE} & \textbf{MSE} & \textbf{MAE} \\
        \midrule 
SOFA Cardiovascular (Max, 24hr) & $\mathbf{0.63 \pm 0.203}$ & $\mathbf{0.41 \pm 0.081}$ & $0.66 \pm 0.399$ & $0.42 \pm 0.141$ \\ 
SOFA CNS (Max, 24hr) & $\mathbf{0.13 \pm 0.032}$ & $\mathbf{0.26 \pm 0.037}$ & $0.38 \pm 0.366$ & $0.39 \pm 0.150$ \\ 
SOFA Renal (Max, 24hr) & $\mathbf{0.14 \pm 0.065}$ & $\mathbf{0.27 \pm 0.062}$ & $0.17 \pm 0.085$ & $0.31 \pm 0.098$ \\ 
SOFA Respiration (Max, 24hr) & $0.46 \pm 0.714$ & $\mathbf{0.38 \pm 0.178}$ & $\mathbf{0.36 \pm 0.307}$ & $0.40 \pm 0.155$ \\ 
SOFA Cardiovascular (Avg) & $0.43 \pm 0.220$ & $\mathbf{0.24 \pm 0.059}$ & $\mathbf{0.32 \pm 0.117}$ & $0.27 \pm 0.083$ \\ 
SOFA CNS (Avg) & $0.05 \pm 0.058$ & $\mathbf{0.12 \pm 0.089}$ & $\mathbf{0.04 \pm 0.027}$ & $0.15 \pm 0.059$ \\ 
SOFA Renal (Avg) & $0.43 \pm 0.689$ & $0.34 \pm 0.214$ & $\mathbf{0.10 \pm 0.033}$ & $\mathbf{0.23 \pm 0.069}$ \\ 
SOFA Respiration (Avg) & $0.48 \pm 0.865$ & $0.31 \pm 0.243$ & $\mathbf{0.10 \pm 0.160}$ & $\mathbf{0.20 \pm 0.195}$ \\ 
SOFA Cardiovascular (Max) & $0.88 \pm 0.526$ & $0.37 \pm 0.157$ & $0.88 \pm 0.435$ & $\mathbf{0.35 \pm 0.084}$ \\ 
SOFA CNS (Max) & $\mathbf{0.06 \pm 0.036}$ & $\mathbf{0.14 \pm 0.036}$ & $0.07 \pm 0.034$ & $0.19 \pm 0.074$ \\ 
SOFA Renal (Max) & $\mathbf{0.16 \pm 0.054}$ & $0.29 \pm 0.084$ & $0.18 \pm 0.055$ & $\mathbf{0.28 \pm 0.060}$ \\ 
SOFA Respiration (Max) & $0.09 \pm 0.073$ & $0.23 \pm 0.126$ & $\mathbf{0.07 \pm 0.039}$ & $\mathbf{0.18 \pm 0.040}$ \\ 
        \bottomrule
        \textbf{Concept} & \multicolumn{2}{c}{\textbf{Vanilla CBM}} & \multicolumn{2}{c}{\textbf{Context-Aware CBM}} \\
        \cmidrule(lr){2-3} \cmidrule(lr){4-5}
        & \textbf{Accuracy} & \textbf{Recall} & \textbf{Accuracy} & \textbf{Recall} \\
        \midrule 
Respiratory Comorbidity (Moderate) & $0.85 \pm 0.048$ & $0.96 \pm 0.000$ & $\mathbf{0.89 \pm 0.017}$ & $\mathbf{0.96 \pm 0.015}$ \\ 
Respiratory Comorbidity (Severe) & $0.98 \pm 0.002$ & $0.95 \pm 0.005$ & $\mathbf{0.98 \pm 0.001}$ & $\mathbf{0.96 \pm 0.005}$ \\
        \bottomrule        
    \end{tabular}%
    }
    \caption{Corresponding tabulated results for Figure \ref{fig:cbm_concepts}. Concept prediction performance for Vanilla CBM and Context-Aware CBM using MSE and MAE for regression tasks, and Accuracy and Recall for classification tasks. Better values are highlighted in bold. We observe, Context-aware CBM outperforms Vanilla CBMs across all respiration concepts, which is critical for the final classification of ARDS. This shows, augmenting LLM concepts leads to assigning a higher weight to the relevant concepts, which ultimately improves the final prediction, thereby improving mutual information and reducing leakage. Here, bold indicates better.}
    \label{tab:cbm_concepts}
\end{table*}

\begin{figure}[h]
    \centering
\includegraphics[width=\linewidth]{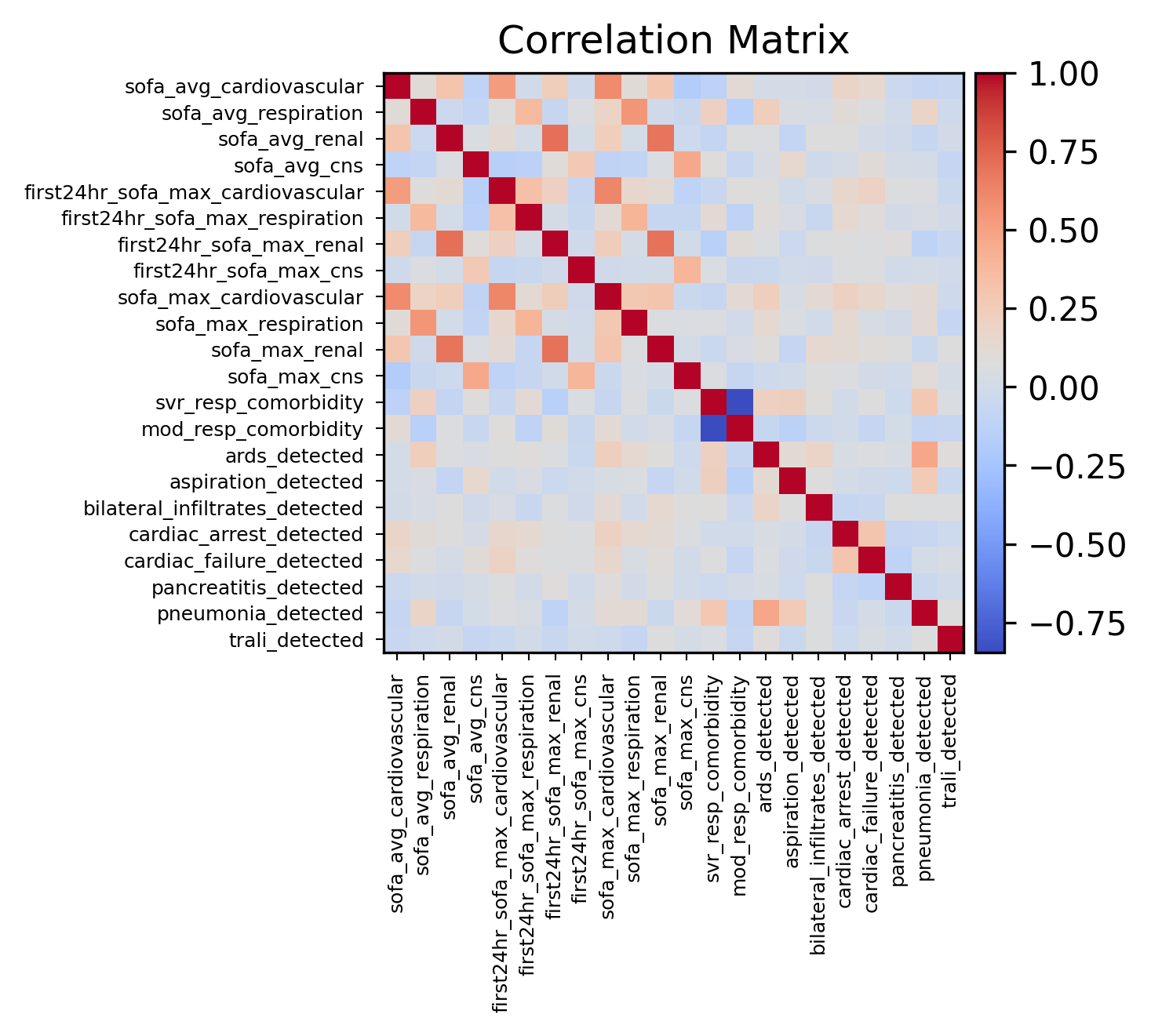}
    \label{fig:correlations}
    \caption{Concept Correlations. We observe strong positive correlations among average cardiovascular SOFA scores, maximum cardiovascular SOFA scores over the first 24 hours, and overall maximum cardiovascular SOFA scores. Similar strong correlations are also observed within the respiration, renal, and CNS-related concepts. Additionally, there is a strong negative correlation between severe and moderate respiratory comorbidities, which aligns with clinical intuition. Among LLM-derived concepts, we find high correlations between impression of ARDS and both pneumonia and bilateral infiltrates, as well as between bilateral infiltrates and pneumonia detection. Because many concepts are strongly correlated, interventions that account for cross-concept dependencies work better than those assuming independence.}
\end{figure}

\begin{figure}[h]
    \centering
    \includegraphics[scale=0.31]{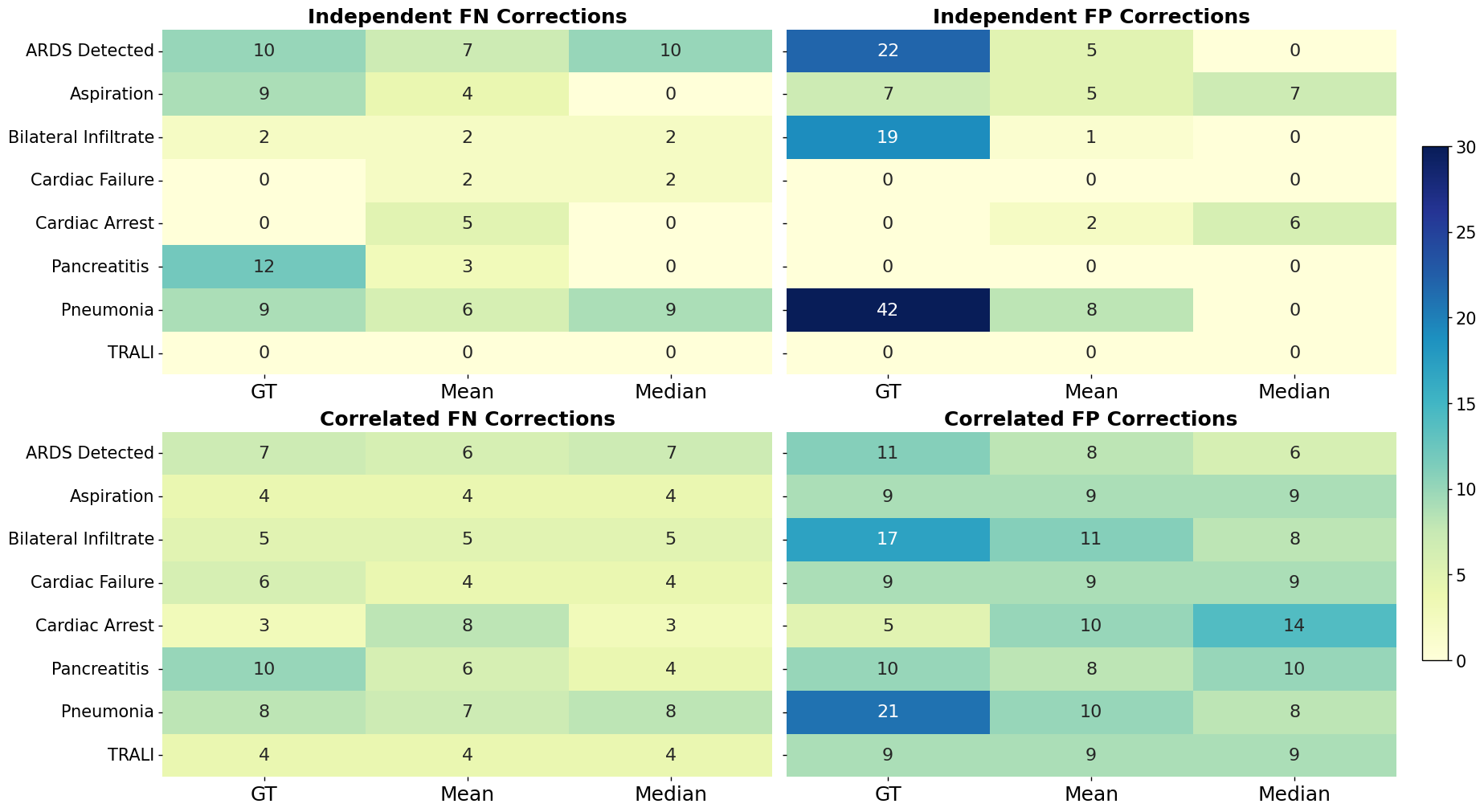}
    \caption{Interventions across individual LLM concepts. See Figure \ref{fig:fp_fn_vanilla_concepts} for vanilla concepts. Similar to the vanilla case, we observe more corrections from correlated interventions than from independent ones. However, in this case, the highest number of corrections comes from interventions based on the ground truth method.}
    \label{fig:fp_fn_concepts}
\end{figure}

\begin{table*}[h]
\centering
\resizebox{\textwidth}{!}{%
\begin{tabular}{p{8cm}ccccc}
\toprule
\textbf{LLM Concept} & \textbf{Acc} & \textbf{Prec} & \textbf{Rec} & \textbf{F1} & \textbf{MI} \\
\midrule
Vanilla (Joint)          & 0.69 & 0.71 & 0.72 & 0.71 & 0.08 \\
Vanilla (Seq)            & 0.68 & 0.69 & 0.74 & 0.71 & 0.08 \\
Context-Aware (Joint)    & 0.79 & 0.76 & 0.88 & 0.81 & 0.21 \\
Context-Aware (Seq)      & 0.79 & 0.77 & 0.87 & 0.82 & 0.21 \\
Context-Aware + Vanilla Intervention (Joint) & 0.85 & 0.88 & 0.86 & 0.90 & 0.78 \\
Context-Aware + Vanilla Intervention (Seq)   & 0.84 & 0.82 & 0.82 & 0.88 & 0.76 \\
Context-Aware + LLM Intervention (Joint)     & 0.94 & 0.92 & 1.00 & 0.93 & 0.92 \\
Context-Aware + LLM Intervention (Seq)       & 0.93 & 0.94 & 0.98 & 0.95 & 0.91 \\
Context-Aware + Vanilla + LLM Inter. (Joint)  & 0.96 & 1.00 & 0.93 & 0.97 & 0.96 \\
Context-Aware + Vanilla + LLM Inter. (Seq) & 1.00 & 0.98 & 0.94 & 0.98 & 0.96 \\
\bottomrule
\end{tabular}%
}
\caption{To test whether label leakage comes from joint training, we compared joint CBMs (concepts and label learned together) with sequential CBMs (learn concepts first, then label). Both strategies have similar performance and MI, suggesting that leakage stems not from joint training but from missing concepts in Vanilla CBMs. Context-Aware CBMs mitigate this by expanding the concept space. Interventions are effective in both settings.}
\label{tab:joint_vs_sequential}
\end{table*}

\end{document}